\documentclass[11pt]{article}

\usepackage{coling}
\usepackage{enumitem}

\newif\ifwithappendix
\withappendixfalse

\newif\ifappendixshown\appendixshownfalse

\usepackage[T1]{fontenc}
\usepackage[utf8]{inputenc}
\usepackage[textsize=tiny]{todonotes}
\usepackage{graphicx}
\usepackage{booktabs}
\usepackage{latexsym}
\usepackage{microtype}

\usepackage{footnote}
\makesavenoteenv{tabular}
\makesavenoteenv{table}
\makesavenoteenv{table*}

\newcommand\minput[1]{%
  \input{#1}%
  \ifhmode\ifnum\lastnodetype=11 \unskip\fi\fi}

\usepackage{hyperref}
\usepackage{xstring}

\newcommand{\noqa}[1]{}
\newcommand{\noqall}[1]{}

\usepackage{amsmath}
\usepackage{listings}

\usepackage{siunitx}
\title{LLMzSzŁ: a comprehensive LLM benchmark for Polish}

\author{\bf
  Krzysztof Jassem \And Michał Ciesiółka \And Filip Graliński \AND
  Piotr Jabłoński \And Jakub Pokrywka \And Marek Kubis \AND Monika
  Jabłońska \And Ryszard Staruch \AND \\
  Adam Mickiewicz University \\
  Center for Artificial Intelligence AMU \\
  \texttt{firstname.lastname@amu.edu.pl} \\}

\usepackage{times}
\usepackage{inconsolata}

\begin{document}

\maketitle
\begin{abstract}

This article introduces the first comprehensive benchmark for the
Polish language at this scale: LLMzSzŁ (LLMs Behind the School Desk).
It is based on a coherent collection of Polish national
exams, including both academic and professional tests extracted from
the archives of the Polish Central Examination Board. It covers 4
types of exams, coming from 154 domains. Altogether, it consists of
almost 19k closed-ended questions. We investigate the performance of
open-source multilingual, English, and Polish LLMs to verify LLMs'
abilities to transfer knowledge between languages. Also, the
correlation between LLMs and humans at model accuracy and exam pass
rate levels is examined. We show
that multilingual LLMs can obtain superior results over monolingual ones; however, monolingual models may be beneficial when model size matters. Our analysis highlight the potential of LLMs in assisting with exam validation, particularly in identifying anomalies or errors in examination tasks.

\end{abstract}

\section{Introduction}

LLMs, as a form of Artificial Intelligence, have been invented to perform useful work in a specific environment defined by humans.
Obviously, the question of how to evaluate LLMs has been raised
since their inception \cite{hendrycks2020measuring}. Similarly, humans
are expected to possess general and specialized knowledge to perform
their work and, again, since the beginning of civilization, human
capabilities have been assessed with general and professional exams.

Humans, however, do not operate in an abstract vacuum. Their
performance is always evaluated within a specific culture, its
traditions, and legal system. Language leaves its stamp as well, not
only because exams are always expressed in a particular language, but
also because it is the lens through which reality is understood. One
does not have to accept the Sapir-Whorf Hypothesis in its entirety to argue
that expected answers are not always translatable between cultures and traditions.

In this paper, we present a new Polish LLM benchmark called LLMzSzŁ (\emph{LLM za
Szkolną Ławą}, \emph{LLMs Behind the School Desk}) representing a collection of Polish national exams.
The characteristic feature of the dataset is the inclusion, in one coherent environment, of school and professional examinations. The incorporation of vocational exams makes it possible to
advance hypotheses on LLMs' specific competences for various professions that include:
\begin{itemize}[nosep]
    \item practical application of knowledge in real professional situations,
    \item real-life task solving,
    \item acquaintance of industry regulations, safety standards and applicable laws,
    \item competence in cutting-edge technology,
    \item communication skills,
    \item work organization.
\end{itemize}
In a way, the benchmark
represents what is expected from an `educated' person to be able to
navigate efficiently in the realities of modern Poland.

The design choices we made while developing LLMzSzŁ distinguish it from other benchmarking efforts that rely on the collection of examination materials such as the MMLU proposed by \citet{hendrycks2020measuring}.
First, instead of randomly searching the web for exams and collecting the results,
we decided to rely on a single, reliable, authoritative source of tasks developed by educational experts.
In our case, it is the collection of exams published by the Polish Central Examination Board (Pol. CKE, Centralna Komisja Egzaminacyjna)\footnote{\url{https://cke.gov.pl/}} which is an institution responsible for the
development of testing materials for nationwide exams such as matriculation.
Having a single source of tasks for the benchmark eliminates the problem of incorporating duplicate samples into the dataset.
The credibility of the Polish Central Examination Board minimizes the
risk of introducing questions of low merit or incorrect answers into the benchmark.
Second, our dataset is stratified into the tiers that vary in the complexity of the tasks: middle school exams, which are usually taken at the age of 15, high school exams that require more elaborate language understanding and reasoning skills
and are attempted at the age of 19, and vocational exams that along with reasoning skills also require domain-specific knowledge.
Third, since the collected exam questions are originally written in Polish, the proposed benchmark can be used to verify if the models that were not deliberately trained for Polish
are capable of solving the tasks by transferring the knowledge from the language of their origin.
Finally, since we track the publication date for each exam,
the proposed benchmark can be used to assess the performance of the model under study while minimizing the impact of data contamination on the evaluation results
by restricting the set of test cases to those that were published after the release of the model.
Furthermore, it enables the study of the correlation between the results achieved by the models and those of human examinees, and draw conclusions on a possible use of LLMs to check the difficulty of the future exams before their publication.

Our main contributions are as follows:
\begin{enumerate}
\item Development of a new, multi-tier LLM benchmark that tracks publication time for all the collected samples.

\item Evaluation of a wide range of LLMs with a special focus on models developed for the Polish language.

\item Detailed analysis of the models' performance with regard to the model size, language, release date, and model type (base vs. instruct).

\item Comparison of human and model performance in the presented tasks.
\end{enumerate}

\section{Related Work}

The capabilities of LLMs are verified by their performance against human-prepared benchmarks. In the pre-LLM era, linguistic benchmarks consisted of ''low-level tasks'', such as: grammatical correctness, sentiment analysis, paraphrase detection, semantic similarity, sentence entailment, natural language inference or coreference resolution \cite{wang2018glue}. With the appearance of LLMs, emerging benchmarks have tended to focus on higher-level skills, such as reasoning \cite{sawada2023arb}, problem-solving \cite{zhang2023m3exam} or human-like conversation \cite{liu2024convbench,bai2024mtbench101}.

A cost-effective way to design benchmarks to evaluate LLMs in combined aspects of knowledge, reasoning, and problem-solving is to build them on existing exams prepared for humans. \citet{hendrycks2020measuring} prepared the MMLU benchmark: 57 tests comprising 15,908 questions from humanities, social sciences, STEM and other subjects, collected from freely accessible sources. Later, MMLU was automatically translated (using, among others, DeepL and GPT-3.5) to produce the m\_mmlu dataset for 35 languages\footnote{https://huggingface.co/datasets/alexandrainst/m\_mmlu}. A similar activity was conducted solely for the Spanish language by \citet{plaza2024spanish} with the use of Azure Translator. In the paper, a detailed analysis was carried out to discover the discrepancies between the performances of multilingual models for English and Spanish. \citet{gema2024mmlu} identify numerous errors in the MMLU benchmark, which shows that there is still room for the development of new MMLU-like benchmarks, with a focus on their ground-truthfulness.

The most recent trend in LLM benchmarking is to develop regional MMLU equivalents from national resources. \citet{son2024korean} organized a dataset derived from Korean exams that covered 45 subjects from humanities and STEM. \citet{yüksel2024turkish} developed a smaller benchmark for Turkish (9 subjects), for which questions were prepared by high school experts.
\citet{li2024chinese} created a dataset that covers four main categories: medicine, law, psychology, and education. Most of the questions were taken from university or college exams, but the legal tests were based on the professional qualification examination.

Medicine and law are the most frequent disciplines in the profession-directed benchmark tests. \cite{pokrywka2024pes} is an exam-based dataset existing for Polish, which uses medical questions from the Polish Board Certification Exam. The collection consists of 297 tests (120 questions each) across a broad spectrum of specialties. Experiments carried out on the dataset have revealed huge progress between GPT-3.5 (no tests passed) and the most recent version of GPT 4 (222 tests passed). The follow-up study \cite{grzybowski2024polishmedicalexamsnew} extends the analysis to include the Medical Final Examination and Dental Final Examination in Polish, along with their human-translated English counterparts, introducing a total of 144 new exams. The cross-lingual analysis reveals that large models (\textasciitilde{}70B parameters) perform similarly in Polish and English. In contrast, smaller models perform best when evaluated in the language of their primary training data.

\section{Dataset}
\subsection{Dataset preparation}
The preparation of our dataset began with examination of the structure of each subject in the selected exams. The main characteristic that we focused on was the number of multiple-choice questions with one correct answer. Subjects with a limited number of such questions were excluded as the additional effort required to process these exams would not be justified by the small number of usable questions they would provide. This selection process resulted in the following categories for the dataset: math, natural sciences, biology, physics, and the Polish language for pre-high school questions, and arts, mechanics (including mining and metallurgy), and agriculture (including forestry) for professional exams.

The tests are published annually by the Polish Central Examination Board, which simplifies the task of finding and downloading PDF files.
All exam questions and their corresponding answer keys are stored in separate files on the CKE website, making it crucial to track which answer document matches which question document.

The textual data were extracted using the PyPDF library without the use of LLMs. Extracting all the necessary data from PDF files proved to be straightforward, but due to extreme inconsistencies in the format of the answer keys, matching questions and answers had to be done manually for every type of exam except the professional exams, which were processed using simple scripts. Many PDF files lacked a text layer, making it impossible to extract the questions without using OCR tools.
A universal method for extracting both questions and answers from PDF files could be a valuable step toward scalable data extraction, allowing the dataset to expand annually.

Both the questions and answers had to be cleaned to remove any anomalies in the data, as well as questions requiring additional resources (such as pictures or charts) that could not be processed. We also chose to exclude task numbers and answer labels from our dataset, representing the correct answers by their index in the answer table. Several data-cleaning techniques were applied (e.g., identifying answers that contained processed text beyond a single question). These steps were crucial in preserving as many questions as possible without discarding them due to detected irregularities.

\subsection{Evaluation harness}

As an evaluation environment, the open source LM Evaluation
Harness framework\footnote{\url{https://github.com/EleutherAI/lm-evaluation-harness}}
was used \cite{eval-harness}. The task configuration was based on the
MMLU configuration; in particular, for each answer, a language
model in question was run to return the probability (likelihood), and the
answer with the highest probability was compared with the gold
answer to calculate the accuracy. The following prompt template was used:

\begin{verbatim}
Przykładowe pytanie egzaminacyjne, test
jednokrotnego wyboru
\end{verbatim}
\textit{Eng. Sample exam question, single-choice test}
\begin{verbatim}
{{question.strip()}}
A. {{answers[0]}}
B. {{answers[1]}}
C. {{answers[2]}}
D. {{answers[3]}}
Prawidłowa odpowiedź:
\end{verbatim}
\textit{Eng. The correct answer:}

\subsection{Dataset availability}

The LLMzSzŁ dataset is available at \url{https://huggingface.co/datasets/amu-cai/llmzszl-dataset}.

\section{Evaluation Results}

Table \ref{fig:resulttable} in Appendix \ref{detailedresults} presents
the evaluation results for all variants of the open-weight models tested: Llama \cite{llama}, Phi \cite{phi}, Qwen \cite{qwen}, Bielik \cite{Bielik7Bv01, Bielik11Bv2a}, Qra \cite{Qra}, polish-gpt2 \cite{polishgpt2}, trurl \cite{trurl}, Mistral \cite{mistral}, Mixtral \cite{mixtral}, WizardLM \cite{wizardlm}, Yi \cite{yi}.  All referenced models in the table are accessible via their respective Hugging Face model cards at https://huggingface.co/mistralai/provider/model-name\footnote{For instance, the model mistralai/Mistral-Large-Instruct-2407 can be accessed at \url{https://huggingface.co/mistralai/Mistral-Large-Instruct-2407}.}. For the rest of the paper we skip provider part when referencing a model.

The rest of this section will consider the accuracy of the model with
respect to its size (\ref{modelsize}), language (\ref{modellanguage}),
release date (\ref{modeldate}), and instruction tuning
(\ref{instruct}). See also
\url{https://huggingface.co/spaces/amu-cai/LLMZSZL_Leaderboard} for a
public leaderboard presenting the results for a number of models.

\subsection{Model size} \label{modelsize}
\begin{figure*}
    \centering
    \includegraphics[width=\textwidth]{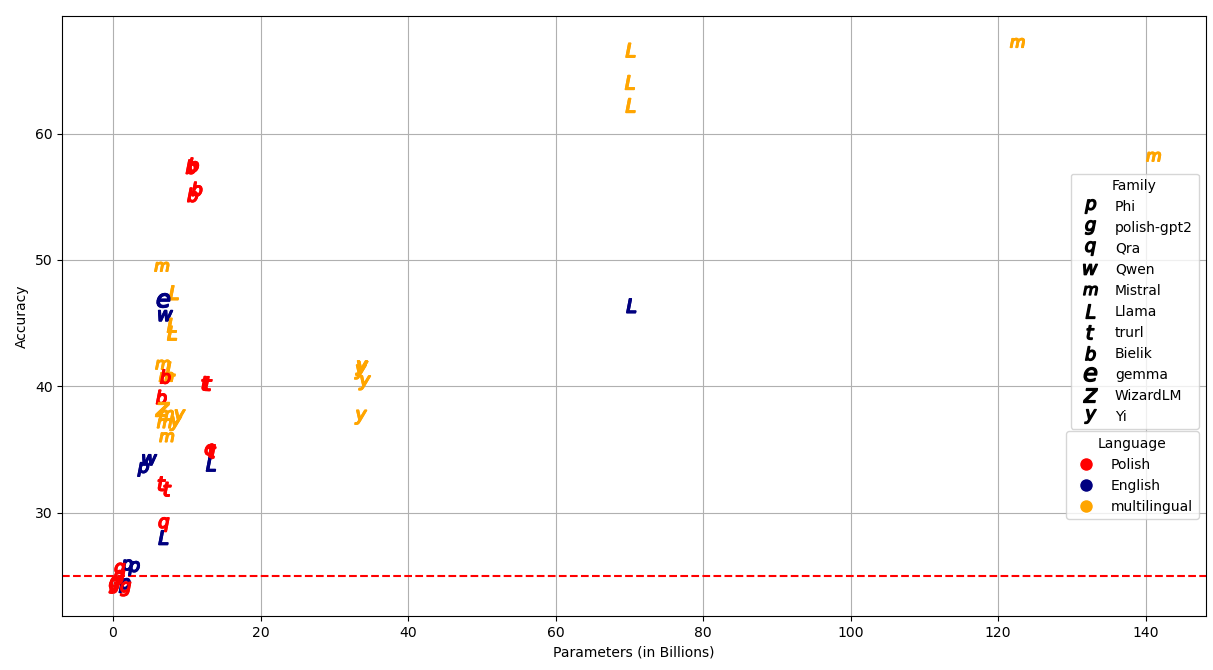}
    \caption{Models' accuracy against their size. The points are jittered in the X-axis for better readability. The red dotted line represents the random guess baseline.}
    \label{fig:scatter}
\end{figure*}

Figure \ref{fig:scatter} illustrates the performance scores plotted against model sizes.
The overall best-performing model is Mistral-Large-Instruct-2407, achieving the accuracy of 67.17. However, the size of the model (123B parameters) may make it difficult to deploy it in certain production environments. The second-best-performing model, Meta-Llama-3.1-70B-Instruct, has a slightly lower accuracy of 66.59 its parameter size is significantly lower (70B). The model still remains impractical for deployment on a single 80GB VRAM GPU in FP16.

Among smaller models (<15B parameters), the best-performing is Bielik-11B-v2.1-Instruct, with an accuracy of 57.52 and a parameter size of 11B. Although its performance is approximately 10 percentage points lower than that of the leading models, it may still be suitable for cost-sensitive production environments. In the category of models with fewer than 8B parameters, notable performers include Meta-Llama-3-8B-Instruct (accuracy 44.83), gemma-7B (accuracy 46.84) and Qwen2-7B (accuracy 45.59). Models smaller than 3B parameters generally perform at the level of random guessing, with an accuracy of approximately 25\%.

The best-performing models within their parameter classes are summarized in Table \ref{tab:bestmodelsparams}.

\begin{table}[h]
\centering
{\small %
\begin{tabular}{crc}
    \hline
    \textbf{Model} & \textbf{Params} & \textbf{Score} \\ \hline
    Mistral-Large-Instruct-2407 & 123B & 67.17 \\
    Meta-Llama-3.1-70B-Instruct & 70B & 66.59 \\
    Bielik-11B-v2.1-Instruct & 11B & 57.52  \\
    gemma-7b & 7B & 46.84  \\
    \hline
\end{tabular}
\caption{Accuracy of the best models in their parameter class}
    \label{tab:bestmodelsparams}
    \centering
}
\end{table}

\subsection{Model language} \label{modellanguage}

When it comes down to models with fewer than 8B parameters, the best performers are either multilingual or English-focused. For slightly larger models, the best performing model is the one fine-tuned for Polish. Overall, the best models are multilingual, though this may be influenced by the lack of models available in Polish and English in this class (except the old Llama-2, which is an older model than Llama-3.1).
Note that not all multilingual models may include Polish in their training datasets. For example, the Yi model family is described as containing English and Chinese, which may explain its low quality of around 40\%, although still above the random guess threshold. We hypothesize that multilingual models, especially large ones, are able to transfer knowledge from other languages to Polish.

\subsection{Release date} \label{modeldate}
\begin{figure*}
    \centering
    \includegraphics[width=\textwidth]{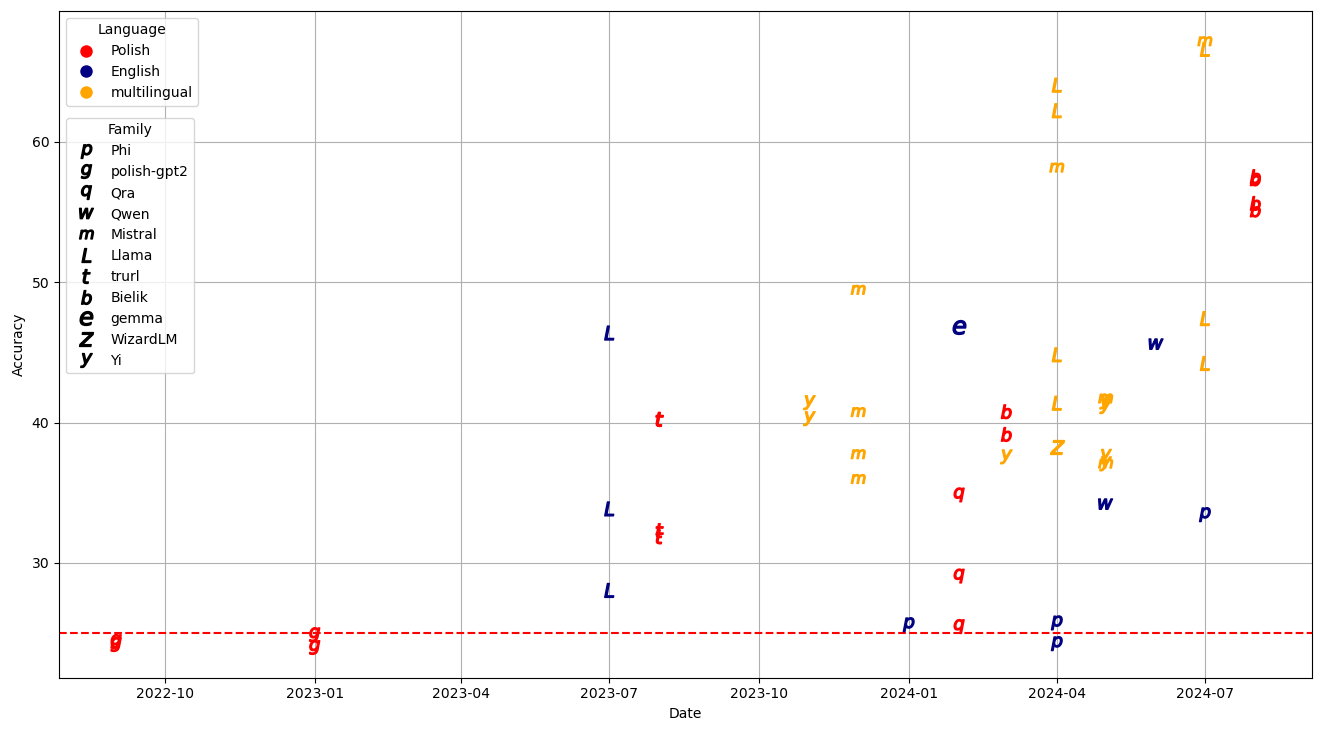}
    \caption{Plot showing the model's accuracy against their release date.}
    \label{fig:scatter2}
\end{figure*}

Figure \ref{fig:scatter2} presents the models' scores against their release date. The first effective models are Llama-2 from July 2023. Since then, the only models that show increasing accuracy across the board have been the Mistral and Llama models.
\subsection{Instruct vs non-instruct models} \label{instruct}

Table \ref{tab:instructgain} describes the difference between a fine-tuned model ('Instruct' or 'Chat' model) and the same model without fine-tuning. For eight out of nine models, instruction tuning improves performance, except for Mistral-7B-v0.1. This leads to the conclusion that fine-tuning generally helps, although it is not always guaranteed.

\begin{table}[h]
\centering
\begin{tabular}{cc}
    \hline
    \textbf{Model} & \textbf{Instruct gain} \\ \hline
     Bielik-7B-v0.1 & \textbf{+1.62}  \\
     Bielik-11B-v2&  \textbf{+0.47} \\
     Meta-Llama-3-8B&  \textbf{+3.45} \\
     Meta-Llama-3-70B& \textbf{+1.82}   \\
     Meta-Llama-3.1-8B& \textbf{+3.20}   \\
     Mistral-7B-v0.1&  -1.77 \\
     Mistral-7B-v0.3& \textbf{+0.67}  \\
     Yi-1.5-9B & \textbf{+0.53} \\
     Yi-1.5-34B & \textbf{+0.33} \\
    \hline
\end{tabular}
\caption{Absolute percentage increase in accuracy: the model fine-tuned on instruction dataset vs not fine-tuned. For example, Meta-Llama-3-8B-Instruct accuracy is 3.45 higher than Meta-Llama-3-8B. Positive values are bolded.}
    \label{tab:instructgain}
    \centering
\end{table}

\section{Detailed evaluation analysis}
\label{sec:detailed}

This analysis aims to find which features of input, expected output, and actual
output are responsible for low results of the models. The Mann-Whitney
U test, which examines the correlation between specific features and the models' scores, can be used for this purpose \cite{gralinski-etal-2019-geval}.
Table~\ref{tab:features} shows a list obtained for the Mistral Large
model. Clearly, answers containing numbers (e.g.
\verb+in<answer1>:SHAPE:9+ meaning a one-digit answer) and, in general, questions related to calculations, pose the greatest challenge for models. This is mirrored by word features,
such as \textit{wynosi} (\textit{is equal}),
\textit{koszt} (\textit{cost}), \textit{oblicz} (\textit{calculate})
or units of measurements (\textit{mm}, \textit{ha}). The most
difficult, for the LLM, were professional exams, in particular,
`Protection and management of forest resources' (code R.13).
Interestingly, the model exhibits some bias, tending to give the wrong
answer if the answer code is B (\texttt{exp:1}).

\begin{table*}[h]
\centering
{\small
\begin{tabular}{llll}
\hline
\textbf{Feature}                             & \textbf{Occurrences} & \textbf{Mean score} & \textbf{Probability}   \\ \hline
in\textless{}answer1\textgreater{}:SHAPE:9   & 1310                 & 1.0215          & 0.000000000000000000 \\
in\textless{}answer2\textgreater{}:SHAPE:9   & 1160                 & 1.0276          & 0.000000000000000000 \\
in\textless{}answer3\textgreater{}:SHAPE:9   & 1061                 & 0.9870          & 0.000000000000000000 \\
in\textless{}question\textgreater{}:SHAPE:9  & 1525                 & 0.9160          & 0.000000000000000000 \\
in\textless{}question\textgreater{}:wynosi   & 1367                 & 0.9140          & 0.000000000000000000 \\
in\textless{}domain\textgreater{}:R.13       & 472                  & 1.1585          & 0.000000000000000000 \\
in\textless{}answer4\textgreater{}:SHAPE:9   & 1009                 & 0.9740          & 0.000000000000000000 \\
in\textless{}question\textgreater{}:Na++podstawie & 227 & 1.2419 & 0.000000000000000000 \\
in\textless{}domain\textgreater{}:M.39       & 391                  & 1.1410          & 0.000000000000000000 \\
exp:1                                        & 4929                 & 0.8748          & 0.000000000000000000 \\
in\textless{}domain\textgreater{}:RL.14      & 185                  & 1.2327          & 0.000000000000000008 \\
in\textless{}domain\textgreater{}:M.11       & 353                  & 1.1052          & 0.000000000000000021 \\
in\textless{}domain\textgreater{}:Matematyka & 365                  & 1.0607          & 0.000000000000007233 \\
in\textless{}answer2\textgreater{}:mm        & 220                  & 1.0661          & 0.000000000000068389 \\
in\textless{}question\textgreater{}:Przedstawiony & 85  & 1.2915 & 0.000000000003937821 \\
in\textless{}question\textgreater{}:koszt    & 252                  & 0.9421          & 0.000000000023219668 \\
in\textless{}question\textgreater{}:oblicz   & 79                   & 1.1537          & 0.000000000779860210 \\
in\textless{}question\textgreater{}:ha       & 162                  & 1.0241          & 0.000000001372934602
\end{tabular}
}
\caption{The features of exam questions correlating with lower scores the most. The
  columns are: feature description, number of occurrences, mean score
  (of cross entropy), probability to obtain the score if a feature did
  not influence the score.}
\label{tab:features}
\end{table*}

An even simpler idea is to analyze the questions for which a model
assigned a very low probability to the expected answer. It might
indicate a mistake on the exam sheet. For example, in this way we
found an error in the 2018 edition of the \textit{Organization and
  management of the maintenance process of motor vehicles} exam
(M.42-X-18.06). The expected answer for Question 24 (\textit{The loyalty
  program is limited to customers only.}) was erroneously specified as
\textit{insolvent} instead of \textit{strategic}.

\section{Evaluating the human-prepared exams by LLMs}

In this section, our objective is to study the correlation between the performances of LLMs and those of human examinees over the years. We acknowledge various limitations of this study. The model scores are restricted to closed questions, whereas the public scores of the examinees include both closed and open questions. In case of professional exams, the public data contain only information on the overall number of passes.

\begin{table*}[h]
\centering
{\small
\begin{tabular}{llllllll}
\hline
\textbf{Type}             & \textbf{Category}              & \textbf{Mistral}        & \textbf{Llama}          & \textbf{Bielik }        & \textbf{gemma} & \textbf{Avg.}  & \textbf{Human} \\ \hline
\multicolumn{2}{l}{Junior High School} & \textbf{43,62} & 34,80          & 32,25          & 28,11 & 34,70 & 47,80 \\
                 & Math                  & \textbf{43,62} & 34,80          & 32,25          & 28,11 & 34,70 & 47,80 \\ \hline
8-grade          &                       & 51,71          & 51,34          & \textbf{55,13} & 33,82 & 48,00 & 55,60 \\
                 & Math                  & \textbf{41,64} & 39,91          & 38,13          & 25,32 & 36,25 & 49,60 \\
                 & Polish                & 61,78          & 62,76          & \textbf{72,12} & 42,33 & 59,75 & 61,60 \\ \hline
High School      &                       & \textbf{60,82} & 58,35          & 44,80          & 34,11 & 49,52 & 35,36 \\
                 & Biology               & \textbf{77,16} & 74,07          & 60,33          & 47,78 & 64,84 & 35,71 \\
                 & Math                  & \textbf{43,81} & 40,20          & 30,99          & 25,43 & 35,11 & 32,44 \\
                 & Physics               & \textbf{67,27} & 67,23          & 47,39          & 31,17 & 53,26 & 39,33 \\ \hline
Professional     &                       & \textbf{62,93} & 61,75          & 56,32          & 38,30 & 54,83 & 82,51 \\
                 & Agri. and Forest.     & 57,93          & \textbf{58,84} & 55,27          & 36,24 & 52,07 & 91,85 \\
                 & Arts                  & \textbf{76,70} & 73,02          & 62,99          & 44,63 & 64,34 & 82,77 \\
                 & MMM                   & \textbf{55,70} & 54,65          & 51,44          & 34,75 & 49,13 & 72,94
\end{tabular}
}
\caption{Average accuracy of best model in each of exams in years 2015-2023. In column "Human" values for Junior, 8-grade, High school exams, are the average achieved score of all examinees in Poland in selected year. For professional exams, "Human" value means percentage of examinees that pass the exam (achieved at least 50.00). Complete tables with results by year are available in the appendix (Tables~\ref{fig:modelperexam_1} and~\ref{fig:modelperexam_2}) }
\label{fig:modelperexam_avg}
\end{table*}

The results obtained by the models were used to verify which language models received the highest scores on certain categories of exams, and what score they received in subsequent years. This made it possible not only to infer which models performed best in a given discipline but, when juxtaposed with the percentage of pass rates by examinees (Table \ref{fig:modelperexam_avg}), to determine whether the exams may have been more difficult or easier in subsequent years.

Of the 38 language models studied, 4 models, with the highest percentage of correct answers per type of exam, were selected for evaluation. The models are shown in Table~\ref{tab:bestmodelsparams}.

\begin{table}[h]
\centering
{\small
\begin{tabular}{lllll}
\hline
\textbf{Exam}    & \textbf{Mistral}                                            & \textbf{Llama} & \textbf{Bielik} & \textbf{gemma}  \\ \hline
Junior High      & \phantom{-}\textbf{0,925} & -0,127         & -0,444          & -0,704          \\
8-grade &
  \phantom{-}0,029 &
  \phantom{-}0,227 &
  \phantom{-}\textbf{0,851} &
  -0,481 \\
High School      & -0,293                                                      & -0,313         & -0,575 & -0,473          \\ \hline
Agri and Forest. & \phantom{-}0,048                           & -0,299         & -0,604          & -0,726 \\
MMM &
  \phantom{-}0,637 &
  \phantom{-}0,591 &
  \phantom{-}0,357 &
  \phantom{-}0,737
\end{tabular}
}
\caption{For Pre-High, 8-grade and High school exams table presents correlation of model scores and student score. For vocational exams (Agriculture and Forestry, MMM - Mechanical, Mining and Metallurgical, Arts), the table presents correlation of model scores and the number of
pass rate.}
\label{fig:corelationexams}
\end{table}

\subsection{Middle school exams} \label{juniohighschool}
In the category of middle school exams (which consist of Junior High school exam and 8-grade exam), the Mistral-Large language model performed the best, achieving an average score of 43.62.

In the years 2015-2019\footnote{In 2020, Poland abolished Junior high schools, and returned to education in the form of primary school (8-grade) and high school (4 years)} significant fluctuations can be observed in both human and model performance. The average real results indicate that the exams were relatively stable in 2015 (48.00) and 2016 (49.00). In 2017, there was a decline to 47, and in the following years the results showed more volatility, rising to 52.00 in 2018, and declining to 43.00 in 2019.

The language models showed a similar trend on average (avg. score). They achieved their highest scores in 2017 (41.64), while in 2019 there was a significant drop to 30.29, which is confirmed in the correlation analysis of individual exams in Table \ref{fig:corelationexams}.

\subsection{High school exams} \label{highschool}
For high school exams, we examined how language models performed in answering questions from three disciplines: biology, mathematics, and physics based on exam sheets from 2015-2023.

Of the disciplines analyzed, the highest score of all the models tested was achieved by Mistral-Large, with an average score of 77.16 in biology, 43.81 in math, and 67.27 in physics, respectively.

\subsubsection{Biology}
The high school biology test scores of the students have steadily decreased, from 43 in 2015 to 26 in 2023, with a marked deterioration in 2018 (32) and 2023. Meanwhile, language model scores rose, reaching 76 in 2018 and 77 in 2023. The lack of correlation may be due to the increase in difficulty of open questions.

\subsubsection{Mathematics}
Students' scores in mathematics steadily declined, from 41 in 2015 to 17 in 2023. In contrast, the results of the language models were stable, oscillating between 30 and 38.

\subsubsection{Physics}
The results of the high school physics exams show clear declines in student performance over the years, from 44 in 2015 to 34 in 2020. There was also a downward trend in the language models -- in 2016 they peaked at 71.84, while in 2019 they dropped to 35.87, which coincides with lower student performance.

\subsection{Professionals exams} \label{proexams}

For professional exams, the statistics for human performance considered the pass rate.

The Mistral-Large turned out to be the highest performer in the arts and mechanical-mining and metallurgical professional exams, with scores of 55.70 in the two disciplines, respectively. In agricultural and forest examinations, Llama-3.1-70B performed best, achieving 58.84.

\subsubsection{Arts}
Over the years, professional art exams have become more difficult for both models and those taking the exam, especially since 2020. However, the biggest gap between the models' and the passers' performance appeared in 2023, when the pass rate dropped to zero, while the models achieved a score comparable to previous years. Finally, we have decided not to include Arts in the correlation analysis because of the small number of data.

\subsubsection{Mechanical, Mining and Metallurgical}
Professional exams in Mechanical and Mining and Metallurgy (MMM) became easier for passers over the years, reaching a peak pass rate in 2022. For language models, the results were more stable, with a slight increase between 2019 and 2023.

\subsubsection{Agriculture and Forestry}
Professional exams in agriculture and forestry were generally easy for students, especially since 2017, when the pass rate exceeded 90. Language models coped steadily with the worksheets in successive years, with results in the 48-55 range of correct answers.

\subsection{Final remarks on correlation between model human performance}

Our research has shown that some models show a high correlation with human performance for school tests (e.g. Mistral for Junior High tests or Bielik for 8-grade schools). If this phenomenon is confirmed with more data (possibly including open questions), it will advocate for a possible use of LLMs as a primary tool for the verification of exam questions prior to their publication. If a model performance differs significantly from previous years in either direction, it may indicate one of the following issues:
\begin{itemize}[nosep]
    \item The test is of different difficulty than in previous years.
    \item The questions are not properly randomized.
    \item Some answers may be erroneous (see Section~\ref{sec:detailed} for an example).
\end{itemize}

In the case of vocational exams, the stability of language model scores over the years may suggest that the difficulty of closed questions has remained similar. Thus, it can be assumed that the decrease in student scores might have been affected by other factors, e.g. by the increase of difficulty of open questions.

\section{Conclusion}

In this paper, we present a new benchmark for large language models built on the basis of Polish school and professional exams.
The proposed benchmark relies on the data collected from a credible, nationwide source. It is stratified by difficulty into middle school, high school, and vocational tiers and contains precise time stamp information for all the collected data.
It is the largest and most comprehensive LLM benchmark developed for the Polish language that has been published to date.
We assessed the performance of a wide range of LLMs with regard to the proposed benchmark and confronted them with the results achieved by humans.
The results show that multilingual LLMs outperform monolingual ones and that the use of monolingual models can be justified when size limitations are a concern.
Furthermore, the outcomes suggest that LLMs can streamline the preparation of future exams by identifying errors in exam questions.

\section{Limitations}
As any competence test that is not designed to test the notion of intelligence in isolation, the presented benchmark intermixes to some extent the measurement of the reasoning capabilities of the model with the assessment of its ability to memorize factual knowledge and answer the questions through a (fuzzy) search with regard to the textual data aggregated in the process of model training.

Taking into consideration that the exams are published online, there is also a risk of data contamination.
Although the exam questions are stored separately from the answers, and they are released in the form of PDF files, this issue cannot be neglected.
To mitigate the risk of data contamination we recommend
tracking the performance of the model under study with regard to our benchmark separately for the exam questions that where published before and after the release date of the model.

Being an exam-based dataset the presented benchmark does not cover informal language or spontaneous speech phenomena that arise in the conversational setting.
The extent to which an examinee's proficiency in solving tests generalizes to performance in real-life situations is an open question.
Measuring real-world competence of an LLM
in skills covered by vocational exams
would require the construction of an embodied agent which was not feasible at the time of writing.
Thus, we deliberately refrained from drawing conclusions in this area.

\bibliography{ms}

\appendix

\section{Detailed Results}
\label{detailedresults}

\begin{table*}
\centering
{\small %
\begin{tabular}{lllSlr}
\hline\textbf{Lang} & \textbf{Family} & \textbf{Name} & \textbf{Parameters (B)} & \textbf{Date} & \textbf{Score} \\
\hline
E & Llama & meta-llama/Llama-2-7b-hf  & 7 & 23-07 & 28.04 \\
 & Llama & meta-llama/Llama-2-13b-hf  & 13 & 23-07 & 33.85 \\
 & Llama & meta-llama/Llama-2-70b-hf  & 70 & 23-07 & 46.38 \\
  & Phi & microsoft/phi-1  & 1 & 24-04 & 25.73 \\
 & Phi & microsoft/phi-1\_5  & 1 & 24-04 & 24.25 \\
 & Phi & microsoft/phi-2  & 3 & 24-01 & 25.60 \\
 & Phi & microsoft/Phi-3-mini-4k-instruct  & 4 & 24-07 & 33.44 \\
 & Qwen & Qwen/Qwen2-1.5B  & 5 & 24-05 & 34.19 \\
 & Qwen & Qwen/Qwen2-7B  & 7 & 24-06 & 45.59 \\
 & gemma & google/gemma-7b  & 7 & 24-02 & 46.84 \\
\hline
P & Bielik & speakleash/Bielik-7B-v0.1  & 7 & 24-03 & 39.15 \\
 & Bielik & speakleash/Bielik-7B-Instruct-v0.1  & 7 & 24-03 & 40.77 \\
 & Bielik & speakleash/Bielik-11B-v2  & 11 & 24-08 & 55.14 \\
 & Bielik & speakleash/Bielik-11B-v2.0-Instruct  & 11 & 24-08 & 55.61 \\
  & Bielik & speakleash/Bielik-11B-v2.1-Instruct  & 11 & 24-08 & 57.52 \\
 & Bielik & speakleash/Bielik-11B-v2.2-Instruct  & 11 & 24-08 & 57.36 \\
 & Qra & OPI-PG/Qra-1b  & 1 & 24-02 & 25.47 \\
 & Qra & OPI-PG/Qra-7b  & 7 & 24-02 & 29.07 \\
 & Qra & OPI-PG/Qra-13b  & 13 & 24-02 & 34.85 \\
 & polish-gpt2 & sdadas/polish-gpt2-small  & 0.2& 22-09 & 24.19 \\
 & polish-gpt2 & sdadas/polish-gpt2-medium  & 0.5& 22-09 & 24.40 \\
 & polish-gpt2 & sdadas/polish-gpt2-large  & 0.9& 23-01 & 24.89 \\
 & polish-gpt2 & sdadas/polish-gpt2-xl  & 2 & 23-01 & 23.98 \\
 & trurl & Voicelab/trurl-2-7b-8bit  & 7 & 23-08 & 31.86 \\
 & trurl & Voicelab/trurl-2-7b  & 7 & 23-08 & 32.30 \\
 & trurl & Voicelab/trurl-2-13b  & 13 & 23-08 & 40.22 \\
 & trurl & Voicelab/trurl-2-13b-8bit  & 13 & 23-08 & 40.23 \\
 & trurl & Voicelab/trurl-2-13b-academic  & 13 & 23-98 & 34.89 \\
\hline
m & Llama & meta-llama/Meta-Llama-3-8B  & 8 & 24-04 & 41.38 \\
 & Llama & meta-llama/Meta-Llama-3-8B-Instruct  & 8 & 24-04 & 44.83 \\
 & Llama & meta-llama/Meta-Llama-3-70B  & 70 & 24-04 & 62.22 \\
  & Llama & meta-llama/Meta-Llama-3-70B-Instruct  & 70 & 24-04 & 64.04 \\
 & Llama & meta-llama/Meta-Llama-3.1-8B  & 8 & 24-07 & 44.21 \\
 & Llama & meta-llama/Meta-Llama-3.1-8B-Instruct  & 8 & 24-07 & 47.41 \\
 & Llama & meta-llama/Meta-Llama-3.1-70B-Instruct  & 70 & 24-07 & 66.59 \\
 & Mistral & mistralai/Mistral-7B-v0.1  & 7 & 23-12 & 37.75 \\
 & Mistral & mistralai/Mixtral-8x7B-Instruct-v0.1  & 7 & 23-12 & 49.46 \\
  & Mistral & mistralai/Mixtral-8x22B-Instruct-v0.1  & 141 & 24-04 & 58.17 \\
 & Mistral & mistralai/Mistral-7B-Instruct-v0.1  & 7 & 23-12 & 35.98 \\
 & Mistral & mistralai/Mistral-7B-Instruct-v0.2  & 7 & 23-12 & 40.75 \\
 & Mistral & mistralai/Mistral-7B-v0.3  & 7 & 24-05 & 37.08 \\
 & Mistral & mistralai/Mistral-7B-Instruct-v0.3  & 7 & 24-05 & 41.72 \\
 & Mistral & mistralai/Mistral-Large-Instruct-2407  & 123 & 24-07 & \textbf{67.17} \\
 & WizardLM & lucyknada/microsoft\_WizardLM-2-7B  & 7 & 24-04 & 38.23 \\
 & Yi & 01-ai/Yi-34B-Chat-4bits  & 34 & 23-11 & 40.28 \\
 & Yi & 01-ai/Yi-34B-Chat  & 34 & 23-11 & 41.42 \\
 & Yi & 01-ai/Yi-34B-200K  & 34 & 24-03 & 37.56 \\
  & Yi & 01-ai/Yi-1.5-9B  & 9 & 24-05 & 37.06 \\
 & Yi & 01-ai/Yi-1.5-9B-Chat  & 9 & 24-05 & 37.59 \\
 & Yi & 01-ai/Yi-1.5-34B  & 34 & 24-05 & 41.14 \\
 & Yi & 01-ai/Yi-1.5-34B-Chat  & 34 & 24-05 & 41.47 \\
\hline\end{tabular}
}
\caption{Accuracy on all of the tested models. In the language column, 'E' stands for English, 'P' -- for Polish, and 'm' -- for multilingual. Parameters are given in billions. The model release date is in YY-MM format. The best score is bolded.}
\label{fig:resulttable}
\end{table*}

\begin{table*}
\centering
{\small %
\begin{tabular}{lllllllll}
\textbf{Year} &
  \textbf{Type} &
  \textbf{Category} &
  \textbf{Mistral} &
  \textbf{Llama} &
  \textbf{Bielik} &
  \textbf{gemma} &
  \textbf{Avg.} &
  \textbf{Human*} \\ \hline
2015 &                     &                   & 60.15 & 61.01 & 41.10 & 29.37 & 47.91 & 57.17  \\
     & Junior high school  &                   & 42.57 & 32.74 & 36.80 & 30.70 & 35.70 & 48.00  \\
     &                     & Math              & 42.57 & 32.74 & 36.80 & 30.70 & 35.70 & 48.00  \\
     & High school         &                   & 65.48 & 72.01 & 23.21 & 20.99 & 45.42 & 42.67  \\
     &                     & Biology           & 76.66 & 95.12 & 4.22  & 15.46 & 47.87 & 43.00  \\
     &                     & Math              & 43.06 & 41.08 & 30.78 & 25.62 & 35.13 & 41.00  \\
     &                     & Phisics           & 76.71 & 79.84 & 34.62 & 21.88 & 53.26 & 44.00  \\
     & Professional        &                   & 60.69 & 59.43 & 60.43 & 37.31 & 54.46 & 74.73  \\
     &                     & Agri. and Forest. & 56.57 & 58.95 & 56.97 & 38.14 & 52.65 & 80.26  \\
     &                     & Arts              & 74.19 & 70.22 & 76.78 & 41.58 & 65.69 & 100.00 \\
     &                     & MMM               & 51.32 & 49.13 & 47.53 & 32.20 & 45.05 & 43.93  \\ \hline
2016 &                     &                   & 56.62 & 49.66 & 50.17 & 35.28 & 47.93 & 46.78  \\
     & Junior high school  &                   & 43.45 & 27.65 & 29.70 & 25.46 & 31.57 & 49.00  \\
     &                     & Math              & 43.45 & 27.65 & 29.70 & 25.46 & 31.57 & 49.00  \\
     & High school         &                   & 63.37 & 53.67 & 54.58 & 39.33 & 52.74 & 36.00  \\
     &                     & Biology           & 70.47 & 35.74 & 66.43 & 46.73 & 54.84 & 36.00  \\
     &                     & Math              & 41.56 & 34.47 & 24.25 & 25.88 & 31.54 & 31.00  \\
     &                     & Phisics           & 78.08 & 90.81 & 73.07 & 45.39 & 71.84 & 41.00  \\
     & Professional        &                   & 53.09 & 54.64 & 53.78 & 34.11 & 48.90 & 61.84  \\
     &                     & Agri. and Forest. & 56.70 & 59.37 & 57.94 & 37.38 & 52.85 & 79.64  \\
     &                     & MMM               & 49.48 & 49.91 & 49.62 & 30.84 & 44.96 & 44.04  \\ \hline
2017 &                     &                   & 59.13 & 57.10 & 54.52 & 38.50 & 52.31 & 58.66  \\
     & Junior high school  &                   & 47.28 & 52.30 & 36.43 & 30.58 & 41.64 & 47.00  \\
     &                     & Math              & 47.28 & 52.30 & 36.43 & 30.58 & 41.64 & 47.00  \\
     & High school         &                   & 57.72 & 52.89 & 57.53 & 40.83 & 52.24 & 38.00  \\
     &                     & Biology           & 53.18 & 51.87 & 86.68 & 59.89 & 62.91 & 37.00  \\
     &                     & Math              & 42.77 & 37.35 & 36.53 & 26.77 & 35.85 & 37.00  \\
     &                     & Phisics           & 77.23 & 69.44 & 49.38 & 35.84 & 57.97 & 40.00  \\
     & Professional        &                   & 64.48 & 62.92 & 57.53 & 38.81 & 55.94 & 83.20  \\
     &                     & Agri. and Forest. & 57.05 & 57.88 & 54.50 & 35.92 & 51.34 & 97.68  \\
     &                     & Arts              & 80.55 & 75.19 & 68.94 & 46.27 & 67.74 & 90.84  \\
     &                     & MMM               & 55.85 & 55.71 & 49.16 & 34.23 & 48.74 & 61.07  \\ \hline
2018 &                     &                   & 67.51 & 62.62 & 47.96 & 35.44 & 53.38 & 61.27  \\
     & Junior high school  &                   & 54.98 & 31.09 & 26.77 & 24.25 & 34.27 & 52.00  \\
     &                     & Math              & 54.98 & 31.09 & 26.77 & 24.25 & 34.27 & 52.00  \\
     & High school         &                   & 73.80 & 71.69 & 45.98 & 36.23 & 56.92 & 32.00  \\
     &                     & Biology           &  96.06 & 98.42 & 54.57 & 54.95 & 76.00 & 32.00  \\
     &                     & Math              & 48.58 & 45.75 & 29.22 & 23.63 & 36.79 & 29.00  \\
     &                     & Phisics           & 76.75 & 70.91 & 54.13 & 30.13 & 57.98 & 35.00  \\
     & Professional        &                   & 65.41 & 64.06 & 57.00 & 38.37 & 56.21 & 93.64  \\
     &                     & Agri. and Forest. & 63.26 & 62.62 & 57.12 & 37.58 & 55.15 & 92.37  \\
     &                     & Arts              & 77.37 & 74.96 & 63.79 & 42.80 & 64.73 & 96.25  \\
     &                     & MMM               & 55.59 & 54.60 & 50.10 & 34.72 & 48.75 & 92.29  \\ \hline
2019 &                     &                   & 53.20 & 53.80 & 48.17 & 34.73 & 47.47 & 63.19  \\
     & Junior high school  &                   & 29.80 & 30.21 & 31.56 & 29.59 & 30.29 & 43.00  \\
     &                     & Math              & 29.80 & 30.21 & 31.56 & 29.59 & 30.29 & 43.00  \\
     & 8-grade exam      &                   & 58.52 & 65.97 & 57.86 & 39.19 & 55.39 & 54.00  \\
     &                     & Math              & 37.26 & 49.10 & 43.95 & 27.77 & 39.52 & 45.00  \\
     &                     & Polish            & 79.79 & 82.84 & 71.77 & 50.61 & 71.26 & 63.00  \\
     & High school         &                   & 43.07 & 38.86 & 34.77 & 26.83 & 35.88 & 40.50  \\
     &                     & Math              & 44.79 & 41.87 & 28.42 & 28.48 & 35.89 & 39.00  \\
     &                     & Phisics           & 41.34 & 35.84 & 41.11 & 25.18 & 35.87 & 42.00  \\
     & Professional        &                   & 64.21 & 63.51 & 56.17 & 38.72 & 55.65 & 91.18  \\
     &                     & Agri. and Forest. & 58.96 & 58.68 & 54.54 & 35.85 & 52.01 & 90.66  \\
     &                     & Arts              & 74.95 & 74.03 & 59.60 & 43.37 & 62.99 & 98.57  \\
     &                     & MMM               & 58.72 & 57.83 & 54.38 & 36.95 & 51.97 & 84.30  \\
\end{tabular}
}
\caption{Accuracy of best model in each of exams. In column "Human" values for Junior, 8-grade, High school exams, are the average achieved score of all examinees in Poland in selected year. For professional exams, "Human" value means percentage of examinees that pass the exam (achieved at least 50.00).}
\label{fig:modelperexam_1}
\end{table*}

\begin{table*}
\centering
{\small %
\begin{tabular}{lllllllll}
\textbf{Year} &
  \textbf{Type} &
  \textbf{Category} &
  \textbf{Mistral} &
  \textbf{Llama} &
  \textbf{Bielik} &
  \textbf{gemma} &
  \textbf{Avg.} &
  \textbf{Human*} \\ \hline
2020 &                     &                   & 53.38 & 51.53 & 44.50 & 32.39 & 45.45 & 60.97  \\
     & 8-grade exam       &                   & 40.38 & 36.39 & 35.47 & 28.51 & 35.19 & 52.50  \\
     &                     & Math              & 44.94 & 41.70 & 41.13 & 29.36 & 39.28 & 46.00  \\
     &                     & Polish            & 35.82 & 31.08 & 29.82 & 27.67 & 31.10 & 59.00  \\
     & High school         &                   & 51.31 & 49.99 & 34.34 & 26.93 & 40.64 & 34.00  \\
     &                     & Math              & 49.10 & 43.44 & 36.67 & 25.25 & 38.62 & 34.00  \\
     &                     & Phisics           & 53.52 & 56.55 & 32.01 & 28.61 & 42.67 & 34.00  \\
     & Professional        &                   & 63.44 & 62.65 & 57.29 & 38.62 & 55.50 & 84.60  \\
     &                     & Agri. and Forest. & 57.90 & 58.32 & 55.33 & 36.58 & 52.03 & 94.57  \\
     &                     & Arts              & 73.42 & 71.40 & 60.46 & 43.03 & 62.08 & 100.00 \\
     &                     & MMM               & 58.99 & 58.21 & 56.10 & 36.26 & 52.39 & 59.22  \\\hline
2021 &                     &                   & 61.60 & 55.65 & 53.56 & 40.01 & 52.71 & 62.98  \\
     & 8-grade exam       &                   & 61.91 & 50.79 & 51.74 & 44.52 & 52.24 & 53.50  \\
     &                     & Math              & 47.52 & 34.34 & 35.77 & 23.11 & 35.19 & 47.00  \\
     &                     & Polish            & 76.29 & 67.23 & 67.71 & 65.92 & 69.29 & 60.00  \\
     & High school         &                   & 60.88 & 52.84 & 50.87 & 38.09 & 50.67 & 32.00  \\
     &                     & Biology           & 84.16 & 72.83 & 76.00 & 50.75 & 70.93 & 33.00  \\
     &                     & Math              & 37.61 & 32.85 & 25.75 & 25.43 & 30.41 & 31.00  \\
     & Professional        &                   & 61.88 & 60.77 & 56.56 & 38.29 & 54.38 & 89.95  \\
     &                     & Agri. and Forest. & 58.19 & 59.45 & 55.69 & 35.90 & 52.31 & 93.23  \\
     &                     & Arts              & 70.93 & 67.70 & 62.02 & 43.25 & 60.97 & 100.00 \\
     &                     & MMM               & 56.52 & 55.17 & 51.98 & 35.71 & 49.84 & 76.63  \\ \hline
2022 &                     &                   & 51.47 & 53.93 & 53.69 & 34.85 & 48.49 & 66.82  \\
     & 8-grade exam        &                   & 25.60 & 38.97 & 61.64 & 25.35 & 37.89 & 58.50  \\
     &                     & Math              & 29.80 & 38.53 & 29.16 & 24.03 & 30.38 & 57.00  \\
     &                     & Polish            & 21.40 & 39.41 & 94.12 & 26.67 & 45.40 & 60.00  \\
     & High school         &                   & 60.38 & 56.86 & 44.87 & 38.94 & 50.26 & 38.00  \\
     &                     & Biology           & 78.60 & 71.80 & 52.99 & 53.78 & 64.29 & 43.00  \\
     &                     & Math              & 42.15 & 41.92 & 36.76 & 24.10 & 36.23 & 33.00  \\
     & Professional        &                   & 62.79 & 61.95 & 54.28 & 38.46 & 54.37 & 91.58  \\
     &                     & Agri. and Forest. & 55.63 & 55.33 & 49.60 & 33.33 & 48.47 & 98.22  \\
     &                     & Arts              & 77.06 & 74.75 & 61.76 & 45.80 & 64.85 & 76.53  \\
     &                     & MMM               & 55.68 & 55.76 & 51.47 & 36.24 & 49.79 & 100.00 \\ \hline
2023 &                     &                   & 67.33 & 65.05 & 58.40 & 37.38 & 57.04 & 51.00  \\
     & 8-grade exam        &                   & 72.15 & 64.58 & 68.92 & 31.55 & 59.30 & 59.50  \\
     &                     & Math              & 48.70 & 35.90 & 40.66 & 22.32 & 36.89 & 53.00  \\
     &                     & Polish            & 95.60 & 93.26 & 97.19 & 40.78 & 81.71 & 66.00  \\
     & High school         &                   & 62.82 & 67.92 & 55.98 & 38.32 & 56.26 & 21.50  \\
     &                     & Biology           & 81.00 & 92.74 & 81.43 & 52.91 & 77.02 & 26.00  \\
     &                     & Math              & 44.65 & 43.10 & 30.53 & 23.72 & 35.50 & 17.00  \\
     & Professional        &                   & 67.12 & 63.46 & 52.99 & 40.65 & 56.06 & 65.00  \\
     &                     & Agri. and Forest. & 57.12 & 58.94 & 55.73 & 35.43 & 51.81 & 100.00 \\
     &                     & Arts              & 85.10 & 75.94 & 50.59 & 50.93 & 65.64 & 0.00   \\
     &                     & MMM               & 59.15 & 55.48 & 52.65 & 35.60 & 50.72 & 95.00  \\
\end{tabular}
}
\caption{Continuation of the previous table: accuracy of best model in each of exams}
\label{fig:modelperexam_2}
\end{table*}

\end{document}